\documentclass[conference]{IEEEtran}
\usepackage{cite}
\usepackage{amsmath,amssymb,amsfonts}
\usepackage{algorithmic}
\usepackage{graphicx}
\usepackage{xcolor}
\usepackage[hidelinks]{hyperref}
\usepackage{comment}
\usepackage{caption}

\begin{document}

\title{An Intelligent Robotic and Bio-Digestor Framework for Smart Waste Management}

\author{
    \IEEEauthorblockN{Radhika Khatri, Adit Tewari, Nikhil Sharma, M. B. Srinivas \\
    \IEEEauthorblockA{BITS Pilani Dubai Campus}
{\{f20220116, f20210915, f20220337, mbs\}@dubai.bits-pilani.ac.in}
}}

\maketitle

\begin{abstract}  
Rapid urbanization and continuous population growth have made municipal solid waste management increasingly challenging. These challenges highlight the need for smarter and automated waste management solutions. This paper presents the design and evaluation of an integrated wa    ste management framework that combines two connected systems, a robotic waste segregation module and an optimized bio-digestor. The robotic waste segregation system uses a MyCobot 280 Jetson Nano robotic arm along with YOLOv8 object detection and robot operating system (ROS)-based path planning to identify and sort waste in real time. It classifies waste into four different categories with high precision, reducing the need for manual intervention. After segregation, the biodegradable waste is transferred to a bio-digestor system equipped with multiple sensors. These sensors continuously monitor key parameters, including temperature, pH, pressure, and motor revolutions per minute. The Particle Swarm Optimization (PSO) algorithm, combined with a regression model, is used to dynamically adjust system parameters. This intelligent optimization approach ensures stable operation and maximizes digestion efficiency under varying environmental conditions. System testing under dynamic conditions demonstrates a sorting accuracy of 98\% along with highly efficient biological conversion. The proposed framework offers a scalable, intelligent, and practical solution for modern waste management, making it suitable for both residential and industrial applications.
\end{abstract}


\section{Introduction}  
\label{intro}
The exponential increase in municipal solid waste poses significant environmental, social, and economic challenges for modern urban centers~\cite{rossit2020exact}. As cities expand, traditional waste handling techniques struggle to keep pace with the sheer volume of refuse generated daily, prompting a paradigm shift toward intelligent and automated solutions~\cite{ghahramani2021iot}. Smart city initiatives increasingly rely on integrating artificial intelligence (AI) and the Internet of Things (IoT) to enhance the efficiency of urban infrastructure~\cite{sosunova2022waste}. However, while considerable progress has been made in optimizing collection routes and monitoring bin capacities, the physical processing and biological treatment of waste at local levels remain highly manual and inefficient. Bridging this gap requires the development of localized, intelligent hardware systems which are capable of both segregating mixed waste streams and processing organic matter autonomously.

Despite recent technological advancements, existing waste management approaches remain insufficient for localized, end-to-end processing for several reasons. First, the majority of contemporary IoT-based waste systems focus almost exclusively on passive monitoring, such as using ultrasonic sensors to detect bin fill levels or AI to optimize collection truck routing~\cite{qurashi2025smart,paul2024iot}. These systems lack active, on-site mechanical intervention, resulting in waste remains mixed and requiring costly downstream sorting. Second, the current biological treatment methods, such as traditional composting or anaerobic digestion, often rely on static operational parameters. They lack real-time algorithmic optimization, leaving them vulnerable to environmental fluctuations and resulting in sub-optimal conversion rates and system instability.

The primary objective of this work is to address these waste management challenges through the development of a highly integrated and automated framework. This work encompasses the design, deployment, and testing of a dual-component system tailored for residential societies and industrial settings. By combining precision robotics with advanced algorithmic optimization for biological degradation, this work aims to minimize the volume of waste destined for landfills while maximizing the yield of usable organic by-products. This approach shifts the focus from mere logistical management to active, on-site material recovery and conversion. To this end, this paper presents a comprehensive system with various contributions. First, we develop an automated robotic waste segregation system using the MyCobot 280 Jetson Nano robotic arm, YOLOv8 computer vision, and ROS-based kinematics to enable real-time sorting of waste into four categories. Second, we propose an optimized bio-digester framework that utilizes real-time sensor data and a Particle Swarm Optimization (PSO)~\cite{erdogmus2018particle} algorithm, supported by a regression model, to dynamically adjust environmental conditions and improve biodegradable waste conversion efficiency.

The remainder of this paper is organized as follows. Section~\ref{rel_work} presents the related work. Section~\ref{prop_sys} describes the proposed system and its components. Section~\ref{methods} details the experimental methodology. Section~\ref{res_dis} discusses the results. Section~\ref{con_fut} concludes the paper with future research directions.

\section{Related Work} 
\label{rel_work}
This section reviews the literature relevant to this work.

\subsection{IoT-Enabled Waste Monitoring Systems} 
The integration of IoT technologies into waste management has been heavily explored to improve municipal collection. The core idea behind these systems involves deploying embedded sensors, such as ultrasonic level detectors and gas sensors, to continuously monitor the physical status of waste bins~\cite{qurashi2025smart,paul2024iot}. These platforms transmit real-time data to centralized cloud servers, enabling municipalities to track accumulation and prevent overflow~\cite{ghahramani2021iot}. The primary strength of this category is its ability to drastically reduce unnecessary collection trips, thereby saving fuel and lowering carbon emissions~\cite{molfese2023optimization}. Transfer learning~\cite{jaiswal2023location,jaiswal2025leveraging,jaiswal2025data} has also been used to design a smart waste management system~\cite{jadli2020toward}. However, a major weakness is that these systems are fundamentally passive, i.e., they monitor the waste but perform no mechanical sorting or processing. In contrast, our proposed framework moves beyond passive observation by actively manipulating and segregating the waste using automated robotics.

\subsection{Computer Vision for Urban Waste} 
One of the prominent areas of research focuses on applying deep learning~\cite{jaiswal2023caqoe} and computer vision~\cite{hewagamage2021computer} to identify and categorize waste in urban environments. Recent studies have developed comprehensive datasets, capturing high-quality images from garbage trucks or street views to train convolutional neural networks for object detection and semantic segmentation~\cite{paulo2026streetview}. The strength of these visual recognition models lies in their high accuracy and ability to map municipal assets dynamically across city streets. Unfortunately, these vision models are frequently deployed in decontextualized environments or strictly for logistical mapping, without being coupled to active sorting hardware. Our work bridges this divide by embedding the state-of-the-art YOLOv8 object detection model directly into the control loop of a robotic manipulator, enabling immediate, physical segregation based on the visual data.

\subsection{Robotics and Optimization in Waste Logistics} 
The application of robotics and mathematical optimization represents the frontier of advanced waste management. A bio-inspired swarm robotics is proposed~\cite{alfeo2019urban} to autonomously forage and collect urban garbage, demonstrating high spatial efficiency. Heuristic algorithms are used to optimize multi-objective accumulation points and sequence collection micro-routes~\cite{rossit2020exact,molfese2023optimization}. While these methods excel at the macro-logistical level, they generally overlook the micro-level biological processing of the collected organic matter. To address them, the framework detailed in this paper complements these logistical optimizations by focusing on the physical processing phase, utilizing a robotic arm for immediate material separation and a Particle Swarm Optimization~ (PSO)~\cite{erdogmus2018particle} algorithm to mathematically optimize the thermodynamic and chemical parameters of the bio-digestion process.

\section{Proposed System} 
\label{prop_sys}

\subsection{System Overview} 
\begin{figure}[t!]
\centering
\includegraphics[width=\columnwidth,height=6.5cm,keepaspectratio]{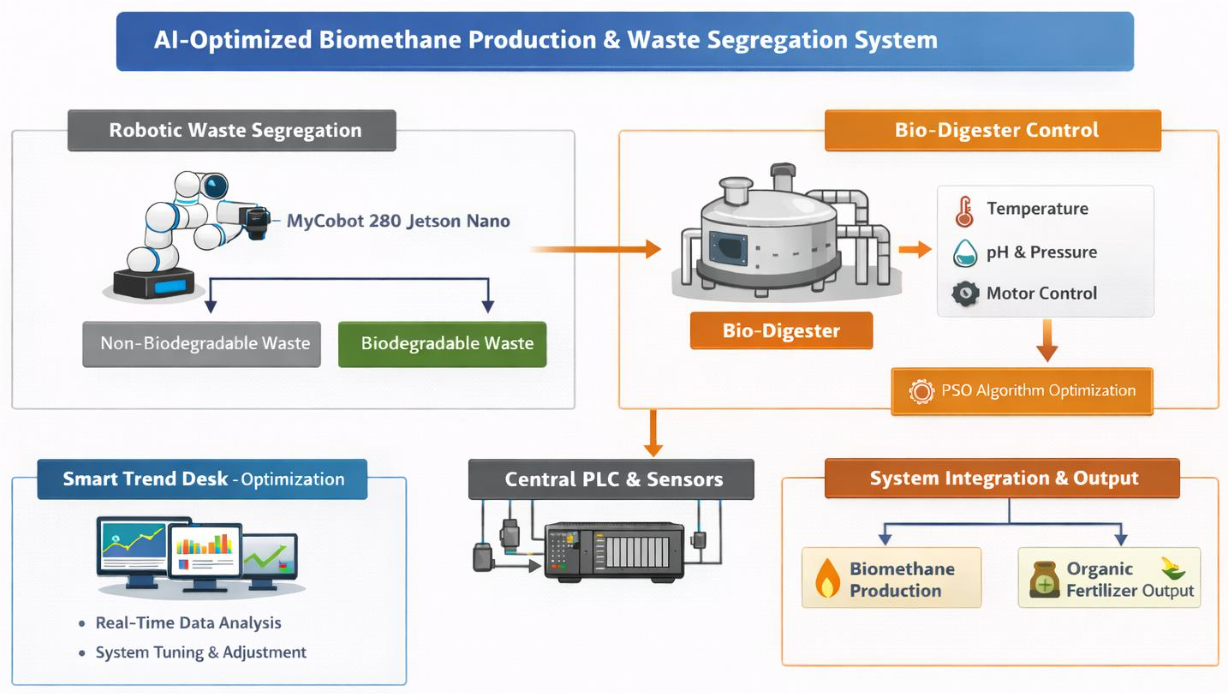}
\caption{The overview of the proposed system.}
\label{pro_sys_over} \vspace{-3mm}
\end{figure}

The proposed system, as shown in Fig.~\ref{pro_sys_over}, is designed as an integrated pipeline that combines perception, actuation, and biological processing into a single automated framework. The process begins with mixed waste being introduced into the system, where a vision module identifies and classifies objects in real time. Based on this classification, the robotic arm executes sorting actions to separate waste into predefined categories. The biodegradable fraction is directed towards a bio-digestor, where it undergoes controlled decomposition. A feedback mechanism continuously monitors system performance and adjusts parameters dynamically. This architecture enables localized waste processing and reduces dependency on centralized waste management facilities.

In addition to its core functionality, the system is designed to operate autonomously with minimal human intervention, making it suitable for real-world deployment in residential and industrial environments. The modular nature of the architecture allows individual components to be upgraded or replaced without affecting the overall system performance. Furthermore, the system is capable of handling varying waste compositions, ensuring adaptability under dynamic conditions. By integrating sensing, computation, and actuation into a unified pipeline, the framework achieves higher efficiency compared to conventional waste management approaches. This holistic design not only improves operational effectiveness but also contributes to environmental sustainability by reducing landfill dependency. Next, we present each component.

\subsection{Robotic Segregation Module} 
\begin{figure}[t!]
\centering
\includegraphics[width=\columnwidth,height=5.5cm,keepaspectratio]{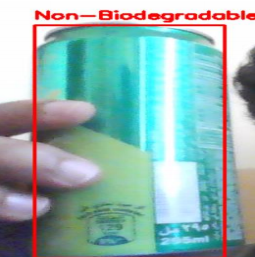}
\caption{Detection of non-biodegradable waste.}
\label{det_bio_waste} 
\end{figure}

The robotic segregation module utilizes the MyCobot 280~\cite{jaksic2025vertical} manipulator, offering flexibility and precision through its six degrees of freedom. A YOLOv8 model~\cite{xiao2024fruit} is employed for object detection and is trained on a dataset representing various waste categories. The detected objects, as shown in Fig.~\ref{det_bio_waste}, are localized in image space and converted into real-world coordinates using calibration techniques. The robot operating system (ROS) framework enables communication between perception and motion planning components, ensuring seamless operation. The trajectory planning algorithms generate efficient and collision-free paths for the robotic arm. This integration allows the system to perform accurate and consistent sorting under varying environmental conditions.

To further enhance performance, the system incorporates real-time feedback mechanisms that validate the success of each pick-and-place operation. Inverse kinematics is used to compute joint configurations required for the precise positioning of the robotic arm. The system is also capable of handling multiple objects sequentially, maintaining a continuous workflow without significant delays. Robustness is achieved through training the detection model on diverse datasets that include variations in lighting, object orientation, and occlusion. This ensures reliable operation even in cluttered environments. Overall, the robotic module significantly reduces manual labor while improving sorting speed and accuracy.

\begin{figure}[t!]
\centering
\includegraphics[width=\columnwidth,height=5.5cm,keepaspectratio]{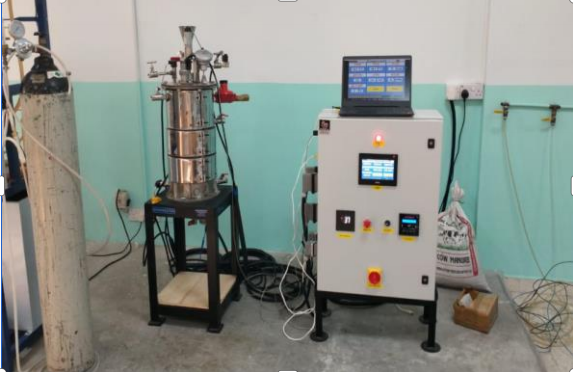}
\caption{Experimental setup of automated bio-digestor system.}
\label{biodig_system} \vspace{-3mm}
\end{figure}

\subsection{Bio-Digestor Optimization Module} 
Fig.~\ref{biodig_system} depicts the bio-digestor module that is responsible for processing organic waste through anaerobic digestion. The sensors continuously monitor environmental parameters such as temperature, pH, and pressure to maintain optimal conditions. A regression model predicts system performance based on sensor inputs, enabling informed decision-making. The Particle Swarm Optimization (PSO) is used to dynamically adjust these parameters for maximum efficiency. The algorithm iteratively refines solutions by balancing exploration and exploitation of the search spaces. This ensures stable operation and improved conversion rates, even under varying input conditions.

\begin{figure}[t!]
\centering
\includegraphics[width=\columnwidth,height=5.5cm,keepaspectratio]{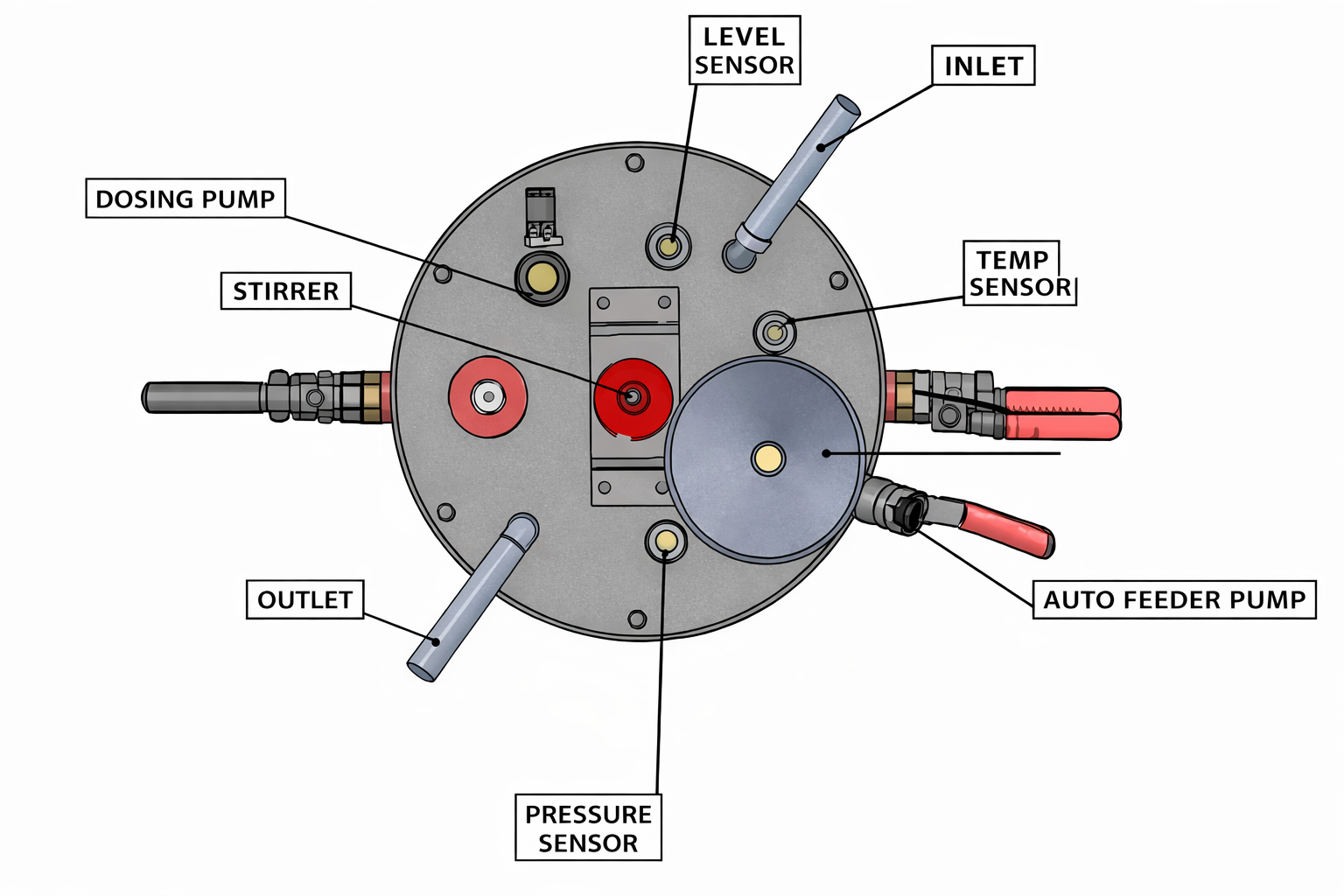}
\caption{Top-view of the bio-digestor.}
\label{biodig_system_part1}
\end{figure}

\begin{figure}[t!]
\centering
\includegraphics[width=\columnwidth,height=5.5cm,keepaspectratio]{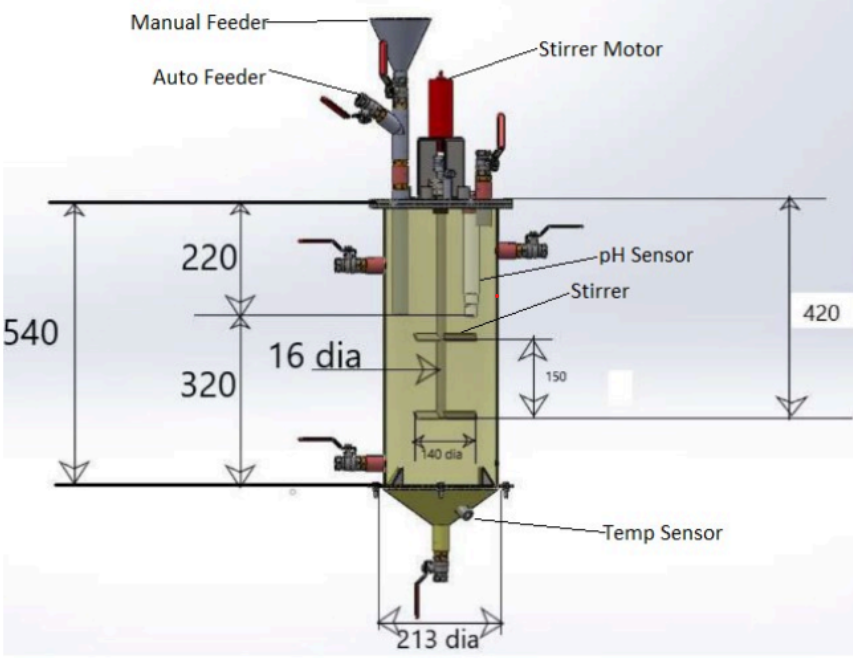}
\caption{Structural design of the bio-digestor system.}
\label{biodig_system_part2} \vspace{-3mm}
\end{figure}

In addition, Fig.~\ref{biodig_system_part1} and Fig.~\ref{biodig_system_part2} present the adaptive digester system designed to handle varying waste compositions and environmental conditions. Continuous monitoring allows early detection of anomalies, preventing system instability and performance degradation. The regression model acts as a predictive layer, estimating the impact of parameter changes before they are applied. This proactive approach minimizes risks and enhances operational reliability. The integration of optimization techniques ensures that the biological process remains efficient over extended periods. As a result, the system achieves higher biogas yield and faster decomposition compared to the traditional static systems.

\subsection{Integration and Control Strategy} 
The integration strategy ensures seamless coordination between the robotic and digestor modules. A centralized controller manages data flow and synchronizes operations across subsystems. The information from the vision module is used to guide robotic actions, while feedback from the digestor informs optimization decisions. This closed-loop system allows real-time adaptation to changes in waste composition and environmental conditions. The modular design also supports scalability and future extensions. Overall, the integration enhances system efficiency and reliability.

Moreover, the system employs a feedback-driven architecture where data from each module contributes to overall decision-making. This interconnected approach ensures that the performance of one module can influence and optimize the operation of others. For example, variations in waste composition detected by the vision system can trigger adjustments in digestor parameters. The centralized control framework also simplifies system management and enables easier debugging and monitoring. By maintaining synchronization across all components, the system achieves consistent and reliable performance. This level of integration is essential for deploying the framework in real-world applications.

\section{Methodology} 
\label{methods}
This section describes the step-by-step methodology employed in the proposed system.

\subsection{Data Acquisition and Preprocessing} 
\label{dataset}
The first stage involves acquiring visual data from the waste input stream. To this end, we used a publicly available waste segregation dataset (augmented) from Kaggle~\cite{waste_dataset_kaggle}. The dataset consists of 56,790 waste images categorized into four classes, namely, food waste, metal, paper, and plastic. The images are in JPG/PNG format with a resolution of 640 × 640 pixels, including both original and augmented samples to improve model robustness. The dataset is divided into training (53,757 images), validation (2,022 images), and testing (1,011 images) to ensure reliable model evaluation. The images are preprocessed to enhance quality and improve detection accuracy.
Techniques such as resizing, normalization, and noise reduction are applied to standardize the input data~\cite{jaiswal2018biometric}. This preprocessing step ensures consistent model performance under varying environmental conditions. Factors such as lighting variations, shadows, and object overlap are addressed during this stage. By improving the quality of input data, the system reduces the likelihood of misclassification and enhances overall reliability. The processed images are then fed to the object detection model for further analysis.

Additionally, data augmentation techniques such as rotation, flipping, and brightness adjustment are used during training to improve model generalization. These techniques help the system handle diverse real-world scenarios effectively. Proper preprocessing ensures that the detection model receives high-quality inputs, which directly impacts overall system accuracy. The combination of hardware placement and software preprocessing creates a strong foundation for subsequent stages.

Moreover, the system generates a large volume of sensor data that must be processed efficiently to support optimization algorithms such as PSO and regression models. Managing these data in real time requires the development of robust data handling and preprocessing pipelines. Initial implementations faced issues related to latency, noise, and data inconsistency. Significant trial and error was involved in selecting appropriate filtering and processing techniques. Ensuring that the processed data is accurate and timely, is essential for effective decision-making. Over time, optimized data pipelines were developed to handle the computational demands of the system.

Integrating multiple sensors, including temperature, pressure, and pH sensors, with the PLC system presented significant initial challenges. Each sensor required careful calibration to ensure accurate and reliable readings under real-time operating conditions. Variations in sensor outputs and environmental noise made it necessary to perform repeated testing and fine-tuning. Ensuring synchronization between different sensors is also critical, as inconsistent data could negatively impact system performance. Additional effort is required to establish stable communication between hardware components and the control system. Through iterative calibration and validation, the system was eventually able to achieve consistent and accurate monitoring.

\subsection{Object Detection and Classification} 
The processed images are analyzed using the YOLOv8 object detection model, which identifies and classifies waste items in real time. The model is trained on the same dataset (see Section \ref{dataset}) containing multiple categories of waste, enabling it to distinguish between biodegradable, recyclable, and hazardous materials. It has high inference speed, allowing the system to process continuous streams of data without significant delay. Each detected object is assigned a bounding box along with a confidence score, indicating the reliability of the classification. This information is used to determine whether the object should be processed further. The model’s performance is optimized through training techniques such as data augmentation and hyperparameter tuning. These enhancements ensure robustness in real-world scenarios where conditions are often unpredictable.

Furthermore, thresholding techniques are applied to filter out low-confidence detections, improving overall system reliability. The model is periodically retrained with new data to adapt to evolving waste patterns. This continuous improvement mechanism ensures long-term effectiveness. The ability to accurately classify waste in real time is a critical component of the system’s success.

\subsection{Coordinate Transformation and Motion Planning} 
Once objects are detected, their positions must be translated into coordinates that the robotic arm can interpret. This is achieved through camera calibration and coordinate transformation techniques. The system maps pixel coordinates to real-world spatial coordinates, enabling accurate positioning of the robotic arm. The robotic control system uses this information to plan its movements. Inverse kinematics is applied to calculate the joint angles required to reach the target position. The trajectory planning algorithms ensure that the movement is efficient, smooth, and free from collisions. This step is crucial for maintaining precision and preventing damage to both the robot and the surrounding environment.

In addition, motion planning algorithms consider constraints such as joint limits and workspace boundaries to ensure safe operation. Optimization techniques are used to minimize movement time and energy consumption, resulting in faster and efficient sorting operations. Accurate transformation and planning are essential to maintain high system performance.

\begin{figure*}[t!]
\centering
\includegraphics[width=\textwidth,height=7.5cm,keepaspectratio]{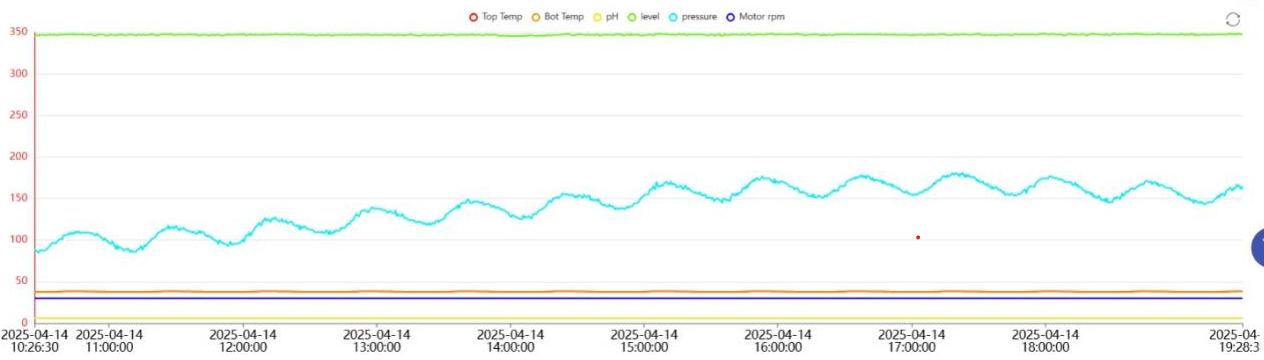}
\caption{Real-time variation of temperature, pH, pressure, and gas production in the biogas chamber during operation.}
\label{para_bio} \vspace{-3mm}
\end{figure*}

\subsection{Robotic Execution} 
In this stage, the robotic arm executes the planned movements to pick and place the detected objects into designated bins. The system ensures that each action is performed with high precision and minimal delay. Feedback mechanisms are used to verify successful execution and make adjustments if necessary. The robotic system is designed to handle multiple objects sequentially, maintaining a continuous workflow. Its performance is evaluated based on metrics such as accuracy, speed, and repeatability. By automating the segregation process, this stage significantly reduces manual effort and improves overall system efficiency.

Error-handling mechanisms are also implemented to manage failed pick attempts or misplacements. The system can reattempt actions or skip objects to maintain workflow continuity. These features enhance robustness and ensure consistent operation. The execution stage is critical for translating computational decisions into physical actions.

\subsection{Bio-Digestor Monitoring and Optimization} 
After segregation, the biodegradable waste is transferred to the bio-digester for further processing. 
The food waste is prepared using two feedstock preparation methods. In the first method, a controlled mixture of bread, cooked rice, milk, and yellow lentils is blended into a homogeneous slurry to maintain a balanced carbon-to-nitrogen ratio suitable for anaerobic digestion. The prepared mixture is further analyzed for Total Solids (TS) and Volatile Solids (VS) using oven drying and ignition methods, and the VS:TS ratio is calculated to evaluate biodegradability and optimize digestion efficiency.
In the second method, lignocellulosic biomass derived from date palm waste is introduced as a secondary substrate. The biomass undergoes alkaline pretreatment using sodium hydroxide (NaOH) to break down lignin and enhance microbial accessibility to cellulose and hemicellulose. After neutralization, washing, and particle size reduction, the prepared slurry is added to the digester to improve methane yield and overall digestion performance.

Sensors continuously collect data on important environmental parameters such as temperature, pH, and pressure. This information helps in monitoring the digestion process and identifying any deviations from optimal operating conditions. A regression model then analyzes the sensor data to predict system performance. Based on these predictions, a Particle Swarm Optimization (PSO) algorithm dynamically adjusts the operational parameters. The algorithm searches for optimal conditions by balancing exploration and exploitation, allowing the digester to maintain efficient performance even when operating conditions change.

Fig.~\ref{para_bio} shows the real-time variation of the key parameters within the biogas chamber during operation. The graph highlights changes in temperature, pH, pressure, and gas production over time. Although minor fluctuations can be observed, the system quickly stabilizes due to the dynamic adjustments made by the PSO algorithm. This continuous monitoring demonstrates the stability of the digestion process and confirms the effectiveness of the optimization strategy in maintaining efficient biological processing.

Furthermore, the system includes safety thresholds to prevent extreme variations in parameters that could negatively impact the biological process. The continuous optimization improves biogas production while reducing overall processing time. The combination of predictive modeling and optimization enables a stable and efficient long-term operation. This stage emphasizes the importance of integrating data-driven techniques with biological systems to enhance performance.

One of the main challenges encountered during development was fine-tuning the PSO algorithm. The optimization process required careful adjustment of parameters such as swarm size, inertia weight, and learning coefficients to achieve stable and efficient performance. Balancing exploration and exploitation was particularly challenging, as inappropriate parameter values sometimes led to instability or slow convergence. Multiple iterations and experimental evaluations were therefore conducted to identify suitable parameter settings. Additionally, integrating the regression model introduced further complexity, since coordination between prediction and optimization components was necessary. Through extensive testing and refinement, a stable configuration was achieved, enabling effective optimization of the digestion process.

\subsection{System Feedback and Adaptation} 
The final stage involves integrating feedback from all modules to enable adaptive system behavior. Data from the robotic system and bio-digestor is continuously analyzed to identify areas for improvement. The system uses this feedback to refine its operations and maintain optimal performance. This adaptive capability is essential for handling real-world variability. It allows the system to respond to changes in waste composition, environmental conditions, and operational constraints. By incorporating feedback into the decision-making process, the methodology ensures long-term reliability and efficiency. Additionally, the feedback system enables predictive maintenance by identifying potential issues before they lead to failure. This reduces downtime and improves system longevity. The adaptive nature of the framework ensures that it remains effective even when the conditions evolve over time. Overall, this stage completes the closed-loop system, making it intelligent, resilient, and scalable.

\section{Results and Discussion} 
\label{res_dis}
This section describes the experimental environment, performance evaluation and obtained results.

\subsection{Experimental Environment} 
The experimental environment consists of an integrated setup combining the robotic waste segregation system and the optimized bio-digestor platform. The waste segregation module utilizes the MyCobot 280 Jetson Nano robotic arm equipped with an adaptive gripper and camera module for object detection and classification. YOLOv8 based vision algorithms and ROS-based motion planning are implemented to enable real-time identification and sorting of waste into four categories, including food waste, metal, paper, and plastic. The bio-digestor system is constructed with a motorized stirring mechanism, and it is embedded with multiple sensors to monitor key operational parameters such as temperature, pH, pressure, level, and motor RPM. These sensors are connected to a PLC-based control and data acquisition system for continuous monitoring and control. A PID-based heating system maintains stable environmental conditions within the chamber, while real-time data logging enables performance tracking and optimization. Additionally, the PSO algorithm and regression model are integrated into the control framework to dynamically adjust operational parameters. The entire setup is tested under varying operational conditions to simulate real-world waste variability and to evaluate system performance and reliability.

\subsection{Performance Evaluation} 
The performance of the proposed system is evaluated using two metrics, one for each major component. For the robotic waste segregation module, classification accuracy~\cite{jaiswal2023caqoe} measures the YOLOv8 model's effectiveness in correctly identifying and categorizing different waste types. The accuracy represents the proportion of correctly classified samples among the total number of predictions. It is defined as~\cite{jaiswal2023caqoe}:

\begin{equation}
\text{Accuracy} = \frac{TP + TN}{TP + TN + FP + FN}
\end{equation}

where $TP$, $TN$, $FP$, and $FN$ represent true positives, true negatives, false positives, and false negatives, respectively. In the context of waste segregation, true positives correspond to correctly identified waste categories, false positives represent incorrectly classified waste items, and false negatives indicate missed detections. These metrics evaluate the overall classification performance of the robotic system.

For the bio-digestor optimization module, the regression model performance is evaluated using the coefficient of determination ($R^2$)~\cite{chicco2021coefficient}. These metrics measure how well the predicted values match the observed values and indicate the accuracy of the regression model in capturing system behavior. It is defined as~\cite{chicco2021coefficient}:

\begin{equation}
R^2 = 1 - \frac{\sum_{i=1}^{n}(y_i - \hat{y}_i)^2}
{\sum_{i=1}^{n}(y_i - \bar{y})^2}
\end{equation}

where $y_i$ represents the observed values, $\hat{y}_i$ represents the predicted values generated by the regression model, and $\bar{y}$ represents the mean of the observed values. The numerator represents the residual sum of squares, while the denominator represents the total sum of squares. The $R^2$ value indicates the proportion of variance in the observed data that is explained by the regression model.

\subsection{Performance Analysis} 
\begin{figure}[b!]
\centering
\includegraphics[width=\columnwidth,height=5.5cm,keepaspectratio]{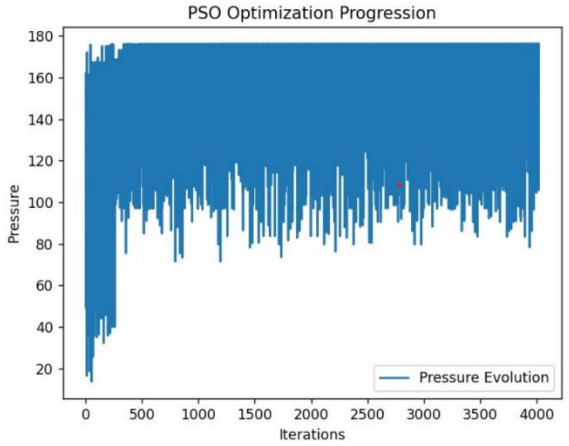}
\caption{Pressure variation during PSO-based optimization.}
\label{outcome_pso} 
\end{figure}

Fig.~\ref{outcome_pso} illustrates the variation in pressure over iterations during the PSO optimisation process. The horizontal axis represents the number of iterations, while the vertical axis indicates the pressure values within the bio-digestor. Initially, fluctuations in pressure are observed as the PSO algorithm explores different parameter combinations. However, as the iterations progress, the pressure gradually stabilizes, indicating convergence towards optimal operating conditions. This behavior demonstrates the effectiveness of the PSO algorithm in achieving stable and optimized system performance.

\begin{figure}[t!]
\centering
\includegraphics[width=\columnwidth,height=5.5cm,keepaspectratio]{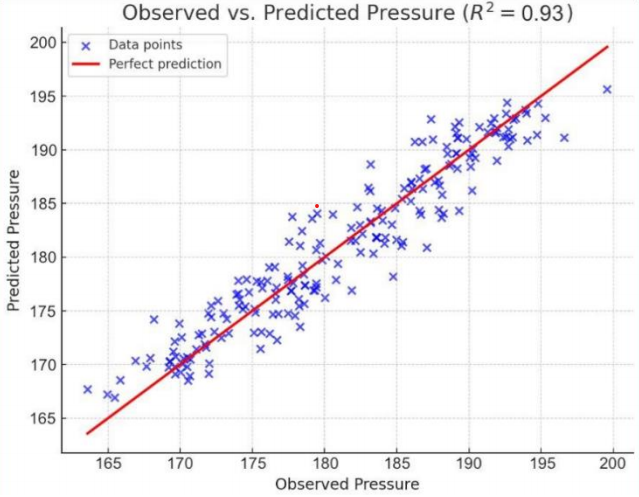}
\caption{The system performance.}
\label{sys_perf} 
\end{figure}

\begin{figure}[t!]
\centering
\includegraphics[width=\columnwidth,height=5.5cm,keepaspectratio]{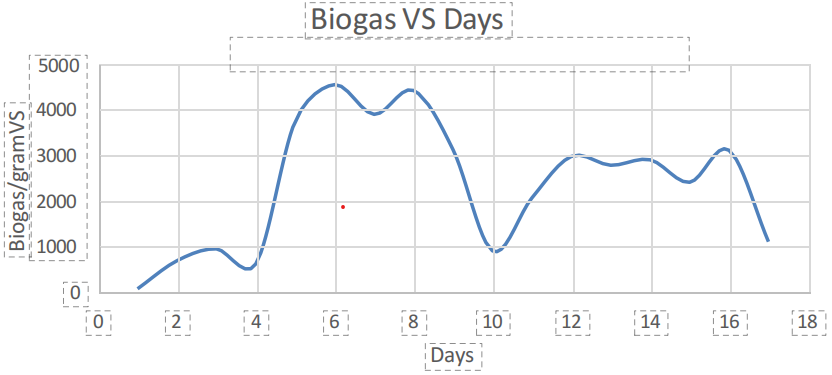}
\caption{Biogas Yield/gramVS over 17 Days.}
\label{foodwaste} 
\end{figure}

\begin{figure}[t!]
\centering
\includegraphics[width=\columnwidth,height=5.5cm,keepaspectratio]{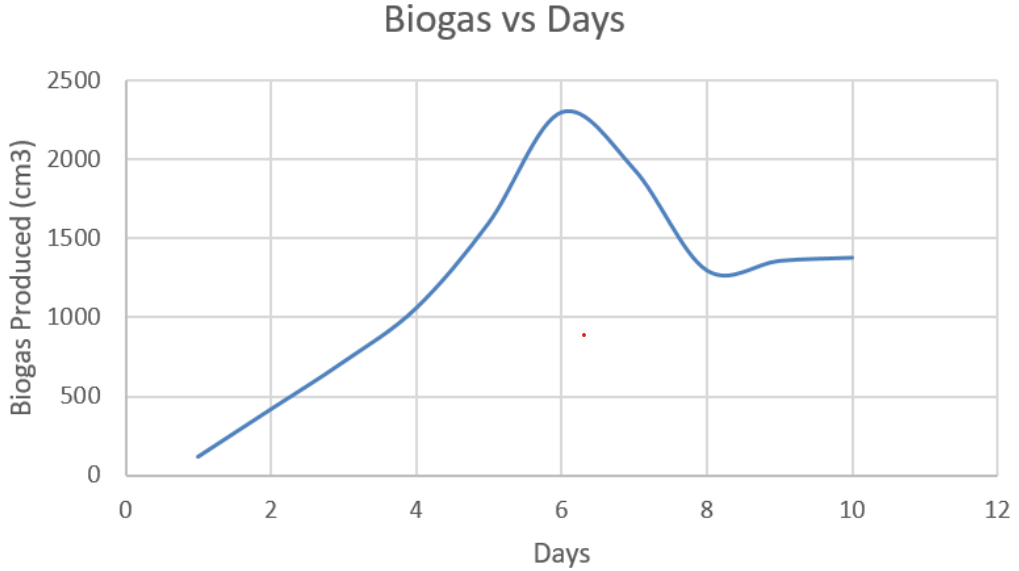}
\caption{Biogas from Lignocellulose (11.1 L in  10 days).}
\label{lignocellulose} \vspace{-3mm}
\end{figure}

Fig.~\ref{sys_perf} presents predicted pressure versus observed pressure obtained from the PSO-based regression model. The horizontal axis represents the observed pressure values measured from the bio-digestor, while the vertical axis represents the predicted pressure values generated by the regression model. The $R^2$ value of 0.93 indicates that the data points closely follow a linear trend, indicating strong agreement between predicted and observed pressure. This relationship demonstrates that the regression model accurately predicts pressure variations and supports effective optimization of the digestion process.

Fig.~\ref{foodwaste} shows the variation of biogas yield per gram of volatile solids (VS) over 17 days. Biogas production initially increases as microbial activity intensifies, reaching peak production around days 5-8. A temporary decline, followed by moderate recovery, indicates substrate depletion and continued digestion of remaining biodegradable material, which demonstrates the dynamic behavior of anaerobic digestion.

Fig.~\ref{lignocellulose} shows biogas production from lignocellulosic (date palm) substrate over 10 days. The biogas production increases steadily, with a peak generation around day 6 due to enhanced biodegradability after alkaline pretreatment. The total biogas production reached approximately 11.1 litres, indicating improved digestion efficiency using lignocellulosic biomass.

Next, the system was evaluated through multiple experimental trials conducted under controlled and semi-realistic conditions. The robotic segregation module achieved an average classification accuracy of 98\% across four waste categories. This accuracy was calculated based on correctly classified objects over the total number of test samples across multiple experimental runs. The experiments were conducted under varying lighting conditions, object orientations, and cluttered environments to assess system robustness. The results demonstrate high reliability and consistent performance in real-time waste segregation tasks.

To summarize, the results suggest that combining robotics with optimization techniques can significantly improve waste management efficiency. The high accuracy of the robotic module demonstrates the effectiveness of integrating YOLOv8 with ROS-based control. Similarly, the performance of the bio-digester highlights the importance of dynamic parameter tuning in biological systems. However, factors such as sensor noise and variations in waste composition may affect performance and need further attention for large-scale deployment. Despite these challenges, our proposed system shows strong potential for practical, real-world applications and sustainable waste management solutions.

\section{Conclusions and Future Work} 
\label{con_fut}
This paper presents an intelligent waste management framework that integrates robotic automation with optimized biological processing. The proposed system efficiently combines a vision-guided robotic segregation module with a sensor-driven bio-digestor, enabling end-to-end, localized waste handling. The YOLOv8 for real-time classification and ROS-based control ensures an accurate and reliable sorting, while the incorporation of PSO allows dynamic tuning of digestor parameters for improved efficiency and stability. Experimental results demonstrate strong performance, achieving high sorting accuracy and consistent biological conversion under varying conditions. This highlights that our proposed framework has the potential to combine robotics, artificial intelligence, and optimization techniques to address modern waste management challenges in a scalable and sustainable manner.

There are several avenues for future research. First, we aim to explore the integration of multi-arm coordination by leveraging swarm robotics principles, enabling multiple robotic manipulators to collaboratively operate on a single conveyor belt to achieve higher throughput. Second, we plan to expand the visual training dataset to include highly distorted, multi-material composite items in order to improve the robustness of the YOLOv8 model under challenging municipal conditions. 


\bibliographystyle{IEEEtran}
\bibliography{references}

\end{document}